\documentclass[11pt]{article}

\usepackage[margin=1in]{geometry}
\usepackage[T1]{fontenc}
\usepackage{lmodern}
\usepackage{microtype}
\usepackage{amsmath,amssymb}
\usepackage{graphicx}
\usepackage{tikz}
\usetikzlibrary{arrows.meta,positioning}
\usepackage{booktabs}
\usepackage{tabularx}
\usepackage{float}
\usepackage{placeins}
\usepackage[colorlinks,urlcolor=blue,linkcolor=blue,citecolor=blue]{hyperref}
\expandafter\def\expandafter\UrlBreaks\expandafter{\UrlBreaks\do\/\do\*\do\-\do\~\do\'\do\"\do\-}

\usepackage{authblk} 

\newcommand{\sptitle}[1]{}         

\newcommand{\jvol}[1]{} \newcommand{\jnum}[1]{} \newcommand{\paper}[1]{}
\newcommand{\jmonth}[1]{} \newcommand{\jname}[1]{} \newcommand{\jtitle}[1]{}

\newcommand{\colrule}{\midrule}    
\newcommand{\botrule}{\hline}

\title{Freshness and the Limits of Heuristic Trend Detection in Temporal RAG}
\author{Matthew Grofsky, D.Eng, CISSP}
\affil{Independent Security Researcher}
\date{}

\begin{document}
\maketitle

\begin{abstract}
Retrieval-augmented generation (RAG) ranks evidence by semantic similarity and ignores time. We
present a lightweight, model-agnostic temporal layer for RAG and use cybersecurity data to separate
two problems that are usually conflated. For \emph{freshness}, surfacing the most recent on-topic
evidence, under corpus-appropriate settings a simple half-life recency prior beats a cosine-only baseline that never
surfaces the newest relevant item (Latest@10~$=0.00$). We test it on three corpora of increasing
realism: a controlled synthetic stream (1.00, but a sanity check, since near-identical intra-topic
embeddings reduce the task to a recency sort), the 849{,}579-event CERT logon corpus (1.00, but only
with corpus-tuned recency weight; 0.00 at the synthetic-tuned default), and a topically-diverse
corpus of NVD CVE descriptions where we verify the freshest item is \emph{not} the most similar
(Latest@10 lifts from 0.00 to 0.60, beating a semantic-then-newest baseline at 0.20). Freshness via a recency prior is thus real but partial and
parameter-sensitive, not solved. For \emph{topic evolution}, labeling weekly clusters as
growth/drift/decay, a fixed-threshold rule scores only 0.08 macro-F1; we localize this failure to
the labeling rule rather than the clusterer (HDBSCAN: 0.10; fixing only the rule: 0.49 in the same
pipeline, 0.96 with clustering noise removed). The contribution is a reproducible decoupling of
freshness from topic evolution, with honest, real-data scope for each, and a reference
implementation.
\end{abstract}

\section{Introduction}{L}arge language models (LLMs) and retrieval-augmented generation (RAG) systems typically rank evidence using only semantic proximity within an embedding space. This \emph{atemporal} view works for timeless facts, but struggles whenever recency, historical validity, or change over time matters. Questions such as \emph{``What was true as of July~2022?''}, \emph{``What changed in the last week?''}, \emph{``What is the latest relevant update?''} require reasoning not only over \emph{what} is similar but also \emph{when} it is relevant. Without a temporal notion of memory, LLM answers are prone to stale evidence, missed updates, and incorrect as-of statements. While vector embeddings excel at capturing semantic similarity, they ignore temporal dynamics. In applied domains, this omission is costly.
\begin{itemize}
    \item In cybersecurity, the attack vector observed today is far more urgent than the same vector observed a year ago.
    \item In sports analytics, the relevance of a player's performance depends on the current season, and not on their career history.
    \item In news and social media, a yesterday's trending topic quickly becomes irrelevant.
\end{itemize}
Without a temporal component, retrieval pipelines risk elevating stale or misleading information. This motivated our proposal: augmenting embeddings using a practical, time-aware scoring method that fuses content similarity with recency weighting.

\textbf{Why time matters.} Embeddings capture \emph{where} items live in a semantic space; however, many answers depend on \emph{when} the evidence is valid. Systems that retrieve only cosine similarity will surface stale context or fail to distinguish novel updates from redundant information (e.g., a vendor's deprecated guidance or last month's roster) \cite{allan01}. In practice, a small, explicit temporal prior can address some freshness failures without retraining.

A useful analogy can be drawn from physics. An event is located in spacetime by both spatial and temporal coordinates; similarly, a document's relevance often depends on its location in semantic space and its timestamp. Therefore, our approach treats retrieval as a joint \textbf{semantic-temporal} problem. Our approach to fusing semantic and temporal scores is inspired by early work in Temporal Information Retrieval, such as Metzler et al. (2009) \cite{metzler09}. However, where their work used a Gaussian distribution to model the temporal focus around a specific event, we adapted this high-level concept for freshness-oriented queries by implementing a fused score that is a convex combination of semantic similarity and a half-life recency prior (Eq.~\eqref{eq:fused}).

We present a lightweight, domain-agnostic \textbf{temporal memory layer} for RAG that makes time a first-class signal throughout the retrieval pipeline. The key idea is simple: preserve timestamps alongside document embeddings and use them in two complementary ways. First, we tracked how topics evolve across time slices via per-slice clustering and one-to-one cross-slice matching, which yields interpretable labels such as \emph{emergence}, \emph{growth}, \emph{decay}, \emph{drift}, and \emph{stable}. Second, at query time we fuse semantic similarity with a recency prior so that newer but still relevant items can outrank older ones when appropriate, while \emph{as-of} queries are answered against a time-bounded snapshot.

\textbf{Research question.} How can retrieval-augmented generation incorporate temporal signals without retraining base models? We propose a simple temporal layer that integrates directly into existing RAG pipelines.

\begin{figure}[H]
  \centering
  \IfFileExists{fig_pipeline.pdf}{%
    \includegraphics[width=\linewidth]{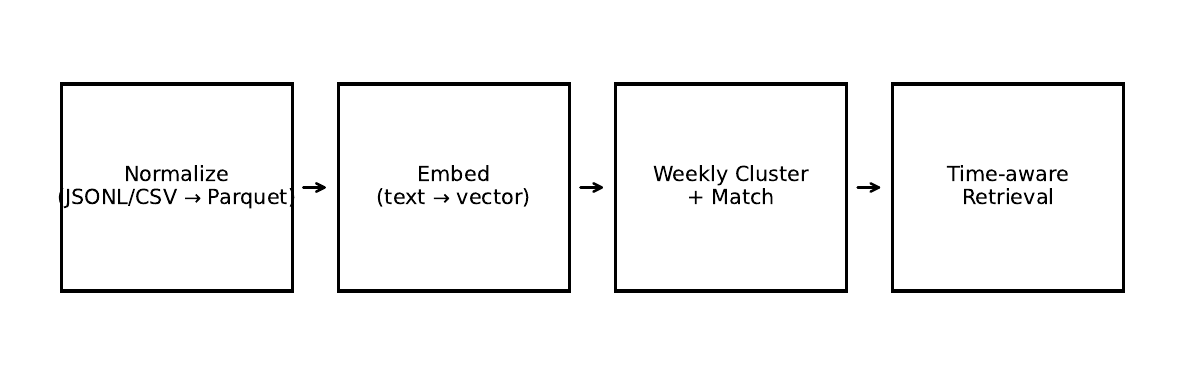}%
  }{%
    \begin{tikzpicture}[node distance=7mm and 8mm,
      box/.style={draw, rounded corners, align=center, minimum height=10mm,
                  minimum width=0.2\linewidth, font=\footnotesize, inner sep=2pt},
      >/.tip={Stealth}]
      \node[box] (n) {Normalize\\\scriptsize JSONL/CSV $\to$ Parquet};
      \node[box, right=of n] (e) {Embed\\\scriptsize text $\to$ vector};
      \node[box, right=of e] (c) {Weekly cluster\\\scriptsize {+} match};
      \node[box, right=of c] (r) {Time-aware\\retrieval};
      \draw[->] (n) -- (e); \draw[->] (e) -- (c); \draw[->] (c) -- (r);
    \end{tikzpicture}%
  }
  \caption{The four-stage pipeline of the temporal memory layer. Raw, unstructured logs are systematically processed into structured artifacts that enable time-aware retrieval.}
  \label{fig:pipeline}
\end{figure}
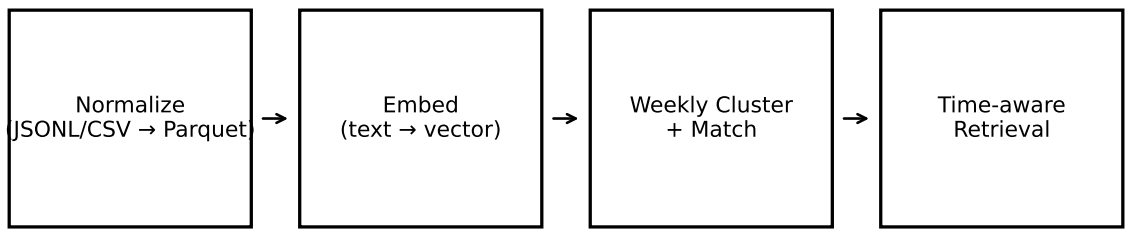

The temporal layer comprises four steps (illustrated in Fig.~\ref{fig:pipeline}).
\begin{enumerate}
    \item \textbf{Data normalization.} Any stream (e.g., security logs, news articles, sports recaps) is mapped to a minimal schema with a UTC timestamp, a compact text representation, and optional domain keys for entities.
    \item \textbf{Embedding.} Text is encoded using a model-agnostic method (e.g., hashing or learned embeddings).
    \item \textbf{Weekly topic tracking.} Items are bucketed by the ISO week (configurable to day/month) and clustered per bucket; clusters are then matched across adjacent buckets using a greedy \emph{one-to-one} assignment on the centroid cosine. Comparing size and similarity to the prior week assigns the labels above; we operationalize semantic change as \emph{drift} $= 1 - \cos(\mathbf{c}_{t-1}, \mathbf{c}_{t})$ and mark drift when it exceeds a small threshold (e.g., $0.2$).
    \item \textbf{Time-aware retrieval.} Given a query vector $q$, document vector $d$, and document time $t$, we re-rank the candidates with a fused score (see Eq.~\eqref{eq:fused}). Optionally, per-slice \emph{cluster summaries} labeled \emph{growth}/\emph{emergence} are boosted inside the user window.
\end{enumerate}

This design is intentionally modest: it requires no retraining of the base model and introduces only a few lines to the existing RAG stacks. However, it addresses three common failure modes.
\begin{itemize}
    \item \emph{First}, \textbf{as-of correctness}: answers conditioned on a date use only evidence $\leq$ that date.
    \item \emph{Second}, \textbf{freshness}: when users implicitly want the latest information, a half-life prior elevates recent, on-topic evidence without discarding older canonical sources.
    \item \emph{Third}, \textbf{change awareness}: the system can surface evolving topics and explain ``what changed'' via growth/decay/drift labels, instead of returning a flat, timeless summary.
\end{itemize}

Although we demonstrate the method with cybersecurity event streams (e.g., vulnerability scans, access patterns, and identity events), the approach is domain-agnostic. News exhibits emerging and waning storylines, sports shows role shifts and injury cycles; product telemetry reveals feature adoption and regressions. In each case, the same four steps, normalize, embed, track per-slice topics, and time-aware re-rank, apply unchanged. The topic labels provide interpretability for operators, whereas the fused score provides a simple, tunable knob for end-user relevance.

\textbf{Scope and positioning.} Our aim was to engineer clarity rather than theoretical novelty. Temporal signals in information retrieval and topic evolution have long been studied. We distill these ideas into a compact, reproducible layer that plugs into modern RAG systems. The method is immediately useful in settings where model retraining is impractical but temporal correctness is important.

\textbf{Paper organization.} Section~\ref{sec:method} details the method and design choices, Section~\ref{sec:impl} outlines the implementation, Section~\ref{sec:rw} discusses related work, Section~\ref{sec:eval} presents the evaluation, Section~\ref{sec:limits} covers the limitations and future directions, and Section~\ref{sec:conclusion} concludes.

\textbf{Contributions.} This study's key contribution is to empirically decouple the problem of temporal RAG into two distinct challenges: a tractable \emph{freshness} problem, which we address with a simple recency prior, and a harder \emph{topic evolution} problem, where we localize the failure of a simple heuristic tracker to its labeling rule rather than its clustering algorithm. We evaluate:
\begin{enumerate}
    \item \textbf{A solution to the \emph{freshness} problem} is to adapt classic temporal re-ranking from Temporal Information Retrieval (TIR). We implement this as a fused semantic-temporal score that blends cosine similarity with a half-life decay function.
    \item \textbf{An investigation of the \emph{topic evolution} problem} was designed to test the performance boundary of simple heuristics. We evaluate a non-probabilistic tracker that uses weekly clustering to assign interpretable labels (\emph{e.g., emergence and drift}). Holding the clustering fixed, replacing the per-week-delta labeling rule with a baseline-relative one raises macro-F1 from 0.08 to 0.49; holding the rule fixed, swapping K-means for HDBSCAN does not help (0.10). The dominant controllable factor is the labeling rule, which motivates change-aware labels (and, for the residual clustering noise, $K$-free methods such as DTMs or HDBSCAN) as future work.
\end{enumerate}
Our results show that the simple recency prior is effective for freshness, while the heuristic tracker's low label score is driven by its rule rather than its clustering, an important correction to the intuition that a more powerful clusterer is the needed fix.

\section{Method}\label{sec:method}
Our goal is to build a \emph{temporal memory layer} that models how topics evolve and answers queries in a time-aware manner, supporting both ``latest'' intent and explicit \emph{as-of} questions. To achieve this in a practical, model-agnostic manner, we followed a four-stage pipeline: (1) normalization, (2) embedding, (3) Weekly Topic Tracking, and (4) Time-Aware Retrieval. This structure is designed for simplicity and can be integrated into any existing RAG stack with a minimal overhead.

\subsection{Stage 1: Data Normalization}
The first stage unifies disparate input formats into a consistent schema. Any incoming data stream, whether from security logs (JSONL) or other sources (CSV), was mapped to a standardized Parquet file. This schema preserves essential temporal and entity information while creating a single, clean input for the subsequent stages.

The core schema fields are:

\vspace{0.5em}
\noindent\begingroup\ttfamily\footnotesize
\{event\_id, ts (UTC), product, event\_type, asset\_id, msg, context(JSON), tech[], attack[], risk\_tag[], text\_repr\}
\par\endgroup
\vspace{0.5em}

Timestamps are coerced to UTC, and a deterministic \texttt{event\_id} is generated if one is not present. The \texttt{text\_repr} field is a compact string concatenation of the most salient fields, designed for effective embedding. This normalization step ensures that the rest of the pipeline is domain-agnostic.

\subsection{Stage 2: Embedding}
Next, the normalized \texttt{text\_repr} for each event is converted into a meaningful numerical vector. To ensure that the vector space was semantically structured for downstream clustering and retrieval tasks, we used a lightweight Sentence-BERT model (\texttt{all-MiniLM-L6-v2}). This model was chosen for its high performance on semantic similarity tasks while maintaining a small footprint suitable for a lightweight pipeline. The resulting 384-dimensional vectors are L2-normalized and stored as \texttt{float16} to balance the precision with the storage efficiency.

\subsection{Stage 3: Weekly Topic Tracking}
To understand how topics evolve, we first grouped the events according to their ISO week. Within each week, we applied K-means clustering to the event vectors to identify dominant themes \cite{manning08}, using it for its simplicity and widespread use as a baseline heuristic. To avoid a fixed K, the optimal number of clusters for each week is determined automatically using the elbow method applied to the within-cluster sum of squares. To create a human-readable summary for each cluster, we extracted the most representative terms using a standard \texttt{CountVectorizer} on the events' text representations.

We employ K-means clustering as a widely understood and computationally efficient baseline. Its known theoretical limitations in high-dimensional, non-spherical spaces make it an ideal candidate for empirically demonstrating the inherent difficulty of the topic evolution task and motivating the need for more advanced methods.

To track the lifecycle of a topic, we matched the clusters between adjacent weeks. A greedy \emph{one-to-one} assignment algorithm links a cluster in week $t$ to its most similar counterpart in week $t-1$ based on the cosine similarity of their centroids. This prevents a single popular topic from matching multiple clusters. Based on this matching, we assigned one of the following trend labels, which are defined by simple, operational rules (defaults in Table~\ref{tab:rules}):
\begin{itemize}
    \item \textbf{Emergence}: A new cluster with no suitable match in the prior week.
    \item \textbf{Growth}: A matched cluster that has significantly increased in size.
    \item \textbf{Decay}: A matched cluster that has significantly decreased in size.
    \item \textbf{Drift}: A matched cluster whose centroid moved, indicating a semantic shift. We defined the drift as $1 - \cos(\mathbf{c}_{t-1}, \mathbf{c}_{t})$, where $\mathbf{c}$ is the cluster centroid.
    \item \textbf{Stable}: A matched cluster that does not meet the criteria for growth, decay, or drift.
\end{itemize}

\begin{table}[!t]
\caption{Trend label rules and defaults.}
\label{tab:rules}
\centering
\begin{tabularx}{\columnwidth}{@{} l >{\raggedright\arraybackslash}X @{}}
\toprule
\textbf{Label} & \textbf{Condition} \\
\colrule
Emergence & No matching prior cluster (sim $< 0.5$) \\
Growth & Size $> 1.5\times$ prior and $\ge$ 30 events \\
Decay  & Size $< 0.5\times$ prior \\
Drift & $1 - \cos(c_{t-1},c_t) \ge 0.2$ \\
Stable & Otherwise \\
\botrule
\end{tabularx}
\end{table}

We note that these rules compare a cluster to its match in the \emph{immediately preceding} week; Section~\ref{sec:eval} shows this choice, rather than the clustering algorithm, is what drives the low trend-label score.

\subsection{Stage 4: Time-Aware Retrieval}
Finally, the retrieval stage uses these artifacts to answer queries. For an \emph{as-of} query, the system first filters its knowledge base to include only documents up to a specified date. For queries implying recency, it re-ranks the top-$K$ semantically similar documents using a fused score that blends semantic relevance with a temporal decay factor as follows:

\begin{equation}
\text{score}(q, d, t) \;=\;
\alpha \,\cos(q,d) \;+\; (1-\alpha)\cdot 0.5^{\,\text{age\_days}(t)/h}
\label{eq:fused}
\end{equation}

Here, $\alpha$ is a tunable weight (default 0.7) that balances cosine similarity with recency, and $h$ is the half-life in days (default 14) that controls how quickly older documents become less relevant. $\text{age\_days}(t)$ is the number of days between a fixed reference time and the document's timestamp $t$; in our evaluation this reference is the most recent timestamp in the corpus (not wall-clock time), matching the implementation, so that results are deterministic across re-runs rather than drifting with the time of execution. This approach ensures that the answers are semantically correct and temporally appropriate. The candidate pool may be a top-$K$ approximate-nearest-neighbour set or the full corpus; our freshness evaluation (Section~\ref{sec:eval}) uses the full pool, because on diverse data the newest relevant item is often not among the top cosine candidates, so a small top-$K$ cutoff would exclude it before re-ranking.

\section{Implementation}\label{sec:impl}
The temporal memory layer was implemented as a series of Python scripts that produced a set of Parquet and CSV artifacts. The pipeline is designed to run sequentially, with each stage building on the previous stage. A Streamlit application was provided for the interactive exploration of the results.

\subsection{Pipeline Stages and Artifacts}
The implementation directly follows the four-stage methodology:
\begin{enumerate}
    \item \textbf{Ingestion (\texttt{log\_ingest.py}):} Raw logs in JSONL or CSV format are read from the \texttt{/logs} directory. The script normalizes records to the standard schema, coerces timestamps to UTC, and writes the output to the \texttt{data/events.parquet}.
    \item \textbf{Embedding (\texttt{embed\_events.py}):} The Parquet file was processed to create a \texttt{text\_repr} for each event. A lightweight SentenceTransformer model then converts these strings into 384-dimensional, L2-normalized semantic vectors, which are saved as \texttt{float16} in \texttt{data/events\_\allowbreak embedded.parquet}.
    \item \textbf{Clustering \& Trends (\texttt{cluster\_trends.py}):} This script groups the embedded events by ISO week, performs K-means clustering, matches clusters to the prior week, and assigns trend labels. The final outputs were two CSV files: \texttt{results/clusters\_weekly.csv} and \texttt{results/trends\_summary.csv}.
    \item \textbf{UI (\texttt{streamlit\_app.py}):} An operator-facing dashboard allows inspection of the results. It features an "as-of" week selector to demonstrate time-scoped filtering, and displays the corresponding trends and cluster summaries.
\end{enumerate}

To regenerate the synthetic and CERT artifacts and evaluation metrics from scratch, run the evaluation scripts:

\begin{verbatim}
python src/eval_metrics_synth.py --force_rerun
python src/eval_metrics_logon.py --force_rerun
\end{verbatim}

\subsection{Parameters and Operational Guidance}
The behavior of the pipeline is controlled by a small set of tunable parameters, exposed as command-line arguments. The thresholds presented in Table~\ref{tab:rules} are empirically derived heuristics chosen to provide a reasonable baseline for detecting meaningful changes in our cybersecurity log dataset. For example, a half-life ($h$) of 14 days was selected to balance the need for freshness with the reality that in many enterprise contexts, information remains highly relevant for several weeks, while a drift threshold of 0.2 was found to effectively filter out minor semantic noise. These values may require tuning of other domains or embedding models. The key parameters are listed in Table~\ref{tab:params}.

\begin{table}[!t]
\caption{Key Operational Parameters and Defaults.}
\label{tab:params}
\centering
\begin{tabularx}{\columnwidth}{@{} l l >{\raggedright\arraybackslash}X @{}}
\toprule
\textbf{Parameter} & \textbf{Default} & \textbf{Purpose} \\
\colrule
\multicolumn{3}{@{}l}{\textit{Weekly Topic Tracking Parameters}} \\
K (clusters) & 6 or auto & Number of weekly topics. \\
Match Threshold & 0.5 & Min cosine sim. for linking clusters. \\
Growth Factor & 1.5 & Min size increase to label as growth. \\
Growth Min. Events & 30 & Min cluster size to qualify for growth. \\
Decay Factor & 0.5 & Max size decrease to label as decay. \\
Drift Threshold & 0.2 & Min $1-\cos$ distance to flag as drift. \\
\multicolumn{3}{@{}l}{\textit{Time-Aware Retrieval Parameters}} \\
Recency $\alpha$ & 0.7 & Weight of semantics vs. recency. \\
Half-life $h$ & 14 days & Time for recency score to halve. \\
\botrule
\end{tabularx}
\end{table}

For operational use, the time slicing granularity can be adjusted; for example, daily slicing may be more appropriate for fast-moving incident response scenarios, whereas default weekly slicing is better for strategic analysis. Similarly,  half-life $h$ should be shortened for domains such as breaking news and lengthened for more stable corpora-like technical documentation.

\subsection{Beyond Security: Sports and News}
\textbf{Sports.} Weekly clustering on play-by-play or injury reports surfaces emerging formations, injuries, or lineup changes; ``as-of'' answers return the last valid roster or scheme at the chosen week.
\textbf{News.} The half-life prior guards against stale facts (positions, prices, and policies), and weekly clusters expose topic drift as a story reframes over time. The same knobs ($\alpha$, $h$) trade recency and stability.

\section{Related Work}\label{sec:rw}

\subsection{Temporal Information Retrieval (TIR)}
Classical Temporal Information Retrieval (TIR) studies how to model time in search, often by filtering results or using temporal priors to favor newer documents~\cite{metzler09,allan01}.
A related challenge is modeling topic evolution, which has been famously addressed using sophisticated generative models.
Dynamic Topic Models (DTMs) use a state-space model to capture the smooth, continuous evolution of topics over time~\cite{blei06dtm}.
We evaluate a simple heuristic-based tracker using discrete-time slices and rule-based labeling as a baseline for this problem space. Where prior framing treats such a tracker's failure as evidence that probabilistic models like DTMs are \emph{necessary}, our experiments (Section~\ref{sec:eval}) show the heuristic tracker's low score is driven by its labeling rule rather than its clustering; we therefore treat richer probabilistic or density-based models as a motivated direction rather than a demonstrated necessity.

\subsection{Temporal Knowledge and Time-Sensitive QA}
Recent studies have focused on enabling AI to answer questions tied to specific times. Temporal Knowledge Graphs (TKGs) do this by adding validity intervals to facts, while benchmarks like TimeQA evaluate an AI's "as-of" correctness. These efforts highlight a critical failure mode in atemporal systems and motivate our goal of providing as-of correct context to an LLM without the need to retrain it.
\subsection{Time-Aware RAG and LLM Systems}
As LLMs have matured, retrieval-augmented systems have begun to incorporate time more explicitly. Some approaches use TKGs within the RAG pipeline, whereas others use freshness-boosting at query time, which is a common practice in production search systems. Our work provides a unified, end-to-end engineering pattern that combines several of these ideas into a single reproducible layer.

\subsection{Positioning}
Our contribution is a practical, lightweight temporal scoring layer, usable for re-ranking when candidate selection is broad or time-aware, that occupies a distinct and valuable position on a spectrum of complexity for time-aware RAG. The current state-of-the-art is not monolithic, but rather a continuum of approaches with varying trade-offs between performance, complexity, and ease of deployment. Recent systems such as TempRALM~\cite{gade2024}, which augments the retriever's scoring function directly, and TA-RAG~\cite{lau2025}, which redesigns the entire pipeline with specialized models, represent more deeply integrated, higher-complexity solutions~\cite{gade2024, lau2025}. In contrast, our approach is a model-agnostic, post-processing step; we note (Section~\ref{sec:eval}) that for freshness it must score a broad or time-aware candidate set, since a narrow top-$K$ cosine cutoff can discard the newest relevant item before the prior acts. Table~\ref{tab:complexity} summarizes these core differentiators, clarifying our method's positioning as a pragmatic, high-impact solution for practitioners for whom model retraining or deep architectural changes are infeasible.

\begin{table*}[!t]
\caption{Spectrum of Complexity in Time-Aware RAG}
\label{tab:complexity}
\centering
\begingroup
\setlength{\tabcolsep}{3pt}
\renewcommand{\arraystretch}{1.15}
\footnotesize
\newcolumntype{Y}{>{\raggedright\arraybackslash}X}
\begin{tabularx}{\textwidth}{Y Y Y c Y}
\toprule
\textbf{Approach} & \textbf{Method} & \textbf{Locus of Change} & \textbf{Cost} & \textbf{System} \\
\colrule
Temporal scoring / broad-candidate re-ranking &
Fused score (semantic + temporal) applied to a broad or time-aware candidate set. &
After retrieval, before generation. &
Minimal &
\textbf{This Work} \\
Retriever Augmentation &
Temporal score is integrated directly into the retriever's ranking function. &
During retrieval. &
Low &
TempRALM~\cite{gade2024} \\
Full Pipeline Redesign &
Temporal query parsing, specialized time-aware retriever models. &
Pre-retrieval \& during retrieval. &
High &
TA-RAG~\cite{lau2025} \\
\botrule
\end{tabularx}
\endgroup
\end{table*}

\section{Evaluation}\label{sec:eval}
We evaluate the temporal memory layer for its ability to correctly identify scripted trends and improve retrieval accuracy for time-sensitive queries across three corpora: a controlled synthetic log stream, the noisy CERT logon dataset, and a topically-diverse corpus of NVD CVE descriptions.

\subsection{Experimental Setup}
Unless otherwise noted, synthetic-data results use the default parameters ($\alpha=0.7$, $h=14$ days). For the real corpora (CERT and NVD~CVE) the optimal recency weight depends on corpus density and timespan, so we report a sweep over $\alpha$ and $h$ and quote the best \emph{observed} setting in that small sweep (CERT $N=2$, NVD $N=5$ queries) rather than a general optimum. Our primary baseline is a standard atemporal \textbf{cosine-only retrieval}, which ranks candidates purely by query--document cosine similarity with no temporal filtering or re-ranking; on the real-data freshness test we additionally compare against a \textbf{semantic-then-newest} baseline (take the top-50 cosine candidates, sort them by timestamp, and evaluate the top 10).

\subsection{Datasets}
\subsubsection{Synthetic Log Stream}
Our first evaluation uses a 12-week stream of synthetic cybersecurity logs with three pre-scripted dynamics, generated using the \texttt{sample\_logs.py} script (seed 42) provided in the supplementary materials~\cite{synthetic-grofsky}. The dataset spans the ISO weeks 2025-W14 to 2025-W26. The Okta authentication failure topic was scripted to \emph{grow} in volume between weeks 4--8. A data access topic was scripted to \emph{drift} semantically from AWS S3 to Snowflake in Week 6. Finally, a vulnerability scanning topic was scripted to \emph{decay} in volume after Week 8. This provides a clear ground truth for measuring the trend detection accuracy.

\subsubsection{Real-World Logon Data}
We also use the public CERT Insider Threat Dataset~\cite{cert-dataset} as a high-volume, real-world corpus to exercise the pipeline at scale. We used the \texttt{logon.csv} file, which contained 849,579 logon and logoff events spanning 71 weeks. We note a data-quality caveat used throughout: every record normalizes to \texttt{product=unknown}, so the text representation is near-constant across events and the cluster vocabulary degenerates. Because the text is near-constant, CERT is a degenerate corpus for trend detection (we report no trend-label F1 on it), but it serves as a high-volume, real-world stress test for the freshness ranking (Section~\ref{sec:eval}, Results on Real-World Data).

\subsubsection{NVD CVE Descriptions}
To test freshness where the newest relevant item is genuinely \emph{not} the most semantically similar, we use a topically-diverse, timestamped, public-domain corpus: CVE records from the National Vulnerability Database~\cite{nvd}. For reproducibility we fetched a fixed snapshot of 4{,}500 CVE records published between 2025-06-29 and 2026-03-05 from the NVD REST API (\texttt{services.nvd.nist.gov/rest/json/cves/2.0}, in 120-day windows) and use their English descriptions and publication dates; the exact corpus (SHA-256 prefix \texttt{b5eb2f90}) is cached as \texttt{data/cve\_corpus.jsonl} and shipped with the repository. Unlike the synthetic and CERT corpora, CVE descriptions carry genuine semantic spread (per-topic cosine standard deviation $0.06$--$0.10$ versus $0.002$ on synthetic), so a query's freshest relevant CVE need not be its most similar; this is the precondition for a meaningful freshness test (Table~\ref{tab:cvespread}).

\begin{table}[!t]
\caption{NVD freshness queries and their deterministic relevance aliases (case-insensitive substring match over CVE descriptions). A CVE counts as relevant to a query if its description contains any listed alias.}
\label{tab:cvequeries}
\centering
\begin{tabularx}{\columnwidth}{@{} >{\raggedright\arraybackslash}X >{\raggedright\arraybackslash}X @{}}
\toprule
\textbf{Query} & \textbf{Relevance alias(es)} \\
\colrule
latest SQL injection vulnerabilities & \texttt{sql injection} \\
recent cross-site scripting (XSS) flaws & \texttt{cross-site scripting}, \texttt{xss} \\
newest remote code execution vulnerabilities & \texttt{remote code execution} \\
recent buffer overflow vulnerabilities & \texttt{buffer overflow} \\
latest privilege escalation vulnerabilities & \texttt{privilege escalation} \\
\botrule
\end{tabularx}
\end{table}

\subsection{Metrics}
We assess performance using three metrics, and state the sample size for each.
\begin{itemize}
    \item \textbf{Trend Label F1 Score:} (Synthetic data only) We measured the macro-F1 score for the classification of weekly topics into label \emph{growth}, \emph{drift}, and \emph{decay}, over \textbf{16 labeled topic-weeks} (growth 5, drift 7, decay 4). We used the macro-average to give equal weight to each class, preventing the common \emph{stable} class from masking poor performance on rare classes; a macro-average over three rare classes and so few points is, however, high-variance, which we account for below.
    \item \textbf{As-of Correctness:} For a query with cutoff date $T$, the knowledge base is first filtered to documents dated $\le T$, and we verify that no returned document post-dates $T$ (pass/fail over 3 queries). Because the candidate pool is pre-filtered on the same timestamp the check inspects, a score of 1.0 confirms the filter is implemented correctly; it is a correctness check on the filter, not an empirical accuracy.
    \item \textbf{Latest@10 / Latest-Set@10:} the fraction of queries for which the newest relevant document appears in the top-10 results. We write \texttt{Latest@10} when the newest item is unique (synthetic and NVD~CVE) and \texttt{Latest-Set@10} when several documents share the terminal timestamp (CERT), counting retrieval of \emph{any} member of the newest set as a success; the two coincide when timestamps are unique. Evaluated over 3 synthetic, 2 CERT, and 5 NVD-CVE topic queries.
\end{itemize}

\subsection{Results on Synthetic Data}
On the synthetic dataset the as-of filter returns 1.00 by construction. For recency queries the fused score places the newest relevant document in the top 10 for every query (Latest@10 = 1.00) while the cosine-only baseline never does (0.00). We stress that this contrast is controlled rather than hard: within each scripted topic the document embeddings are nearly identical, the cosine spread among the 301 Okta-relevant events is 0.002 (Table~\ref{tab:spread}), so similarity carries no ordering signal and the metric reduces to ranking by time. The result confirms the prior is wired correctly; it is not evidence that a hard freshness problem has been solved.

The trend detection model surfaces all three scripted dynamics for operator review. However, the stricter task of assigning \texttt{growth}/\texttt{drift}/\texttt{decay} to the right topic-cluster in the right week scores only 0.08 macro-F1. We trace this to the \emph{labeling rule}, not the clustering. The scripted dynamics are multi-week \emph{states} (an elevated or decayed level that persists), while the rule compares \emph{consecutive} weeks and so can fire only at transition weeks; the growth rule additionally cannot trigger on the scripted magnitude (a $20\to30$ step is not strictly greater than $1.5\times20$). Controlled substitutions isolate the cause (Table~\ref{tab:lever}). Holding the rule fixed and replacing K-means with HDBSCAN leaves the score essentially unchanged (0.10): the choice of clusterer is not the bottleneck. Holding the clustering fixed and replacing the per-week-delta rule with a baseline-relative rule (each week compared to the topic's trailing level) raises macro-F1 from 0.08 to 0.49 in the same pipeline; applying that rule to the topics' true volumes, with clustering set aside, recovers 0.96. The labeling rule is thus the dominant controllable factor, and the residual gap ($0.49\to0.96$) is clustering noise rather than the K-means-versus-density distinction. We therefore report 0.08 as a calibration limit of a specific labeling heuristic, not as a ceiling on clustering for this task.

\begin{table}[!t]
\caption{The lever is the labeling rule, not the clusterer (synthetic, real \texttt{all-MiniLM-L6-v2}).}
\label{tab:lever}
\centering
\begin{tabularx}{\columnwidth}{@{} >{\raggedright\arraybackslash}X c @{}}
\toprule
\textbf{Configuration} & \textbf{Trend macro-F1} \\
\colrule
K-means (elbow) + per-week-delta rule (as published) & 0.08 \\
HDBSCAN + per-week-delta rule (same harness) & 0.10 \\
K-means (elbow) + baseline-relative rule (same pipeline) & 0.49 \\
Baseline-relative rule on true topic volumes (clustering set aside) & \textbf{0.96} \\
\botrule
\end{tabularx}
\end{table}

\begin{table}[!t]
\caption{Why the freshness contrast is trivial here: cosine spread among relevant items.}
\label{tab:spread}
\centering
\begin{tabularx}{\columnwidth}{@{} l c >{\centering\arraybackslash}X c @{}}
\toprule
\textbf{Topic} & \textbf{\# rel.} & \textbf{cosine min / max / std} & \textbf{newest pctile} \\
\colrule
Okta MFA & 301 & 0.692 / 0.697 / 0.002 & 100\% \\
Snowflake/S3 & 309 & 0.380 / 0.512 / 0.056 & 78\% \\
Qualys/OpenSSL & 295 & 0.633 / 0.661 / 0.009 & 57\% \\
\botrule
\end{tabularx}
\end{table}

To characterize the recency prior's operating range, we sweep $\alpha$ (Table~\ref{tab:sensitivity}). The fused score maintains perfect \texttt{Latest@10} for $\alpha \le 0.95$ and degrades only as $\alpha \to 1$; at $\alpha=1.0$ (pure cosine) the newest-relevant rows sit at cosine ranks \#49/\#64/\#124, recovering the 0.00 baseline. The temporal component is thus decisive only in the near-cosine limit; at the default $\alpha=0.7$ the result is not on a knife-edge. (We note that the operating range characterized here supersedes an earlier sweep that reported a breakpoint at $\alpha\!\ge\!0.9$, which we were unable to reproduce.)

\begin{table}[!t]
\caption{Sensitivity of the recency prior ($\alpha$) on synthetic data (regenerated, real model).}
\label{tab:sensitivity}
\centering
\begin{tabularx}{\columnwidth}{@{} l l >{\centering\arraybackslash}X @{}}
\toprule
\textbf{$\alpha$ Value} & \textbf{Description} & \textbf{\texttt{Latest@10} Accuracy} \\
\colrule
0.7 & Default (Balanced) & 1.000 \\
0.9 & Semantic Heavy & 1.000 \\
0.95 & Semantic Dominant & 1.000 \\
0.99 & Near-pure cosine & 0.333 \\
1.0 & Cosine only (baseline) & 0.000 \\
\botrule
\end{tabularx}
\end{table}

\subsection{Results on Real-World Data}
We evaluate freshness on two real corpora; both use the released scripts (\texttt{eval\_cert\_freshness.py}, \texttt{eval\_cve\_freshness.py}) with the reference time set to each corpus's most recent timestamp, and on both the cosine-only baseline never surfaces the newest relevant item (Latest@10 $=0.00$ at every setting).

\textbf{CERT (degenerate real text).} On the 849{,}579-event CERT corpus the fused recency prior recovers the newest event, but only when the recency weight is tuned to this dense, high-volume corpus: at $\alpha=0.3$, $h=3$ it reaches 1.00, whereas at the synthetic-tuned default ($\alpha=0.7$, $h=14$) it collapses to 0.00 (Table~\ref{tab:realfresh}). The synthetic-optimal parameters do not transfer; the recency weight must scale with corpus density and timespan. (We also note, but do not rely on, an earlier observation that the elbow-selected $K$ fluctuated from 3 to 9 week-to-week; given the \texttt{product=unknown} degeneracy this is at most suggestive of $K$-instability.)

\textbf{NVD CVE (diverse real text).} This is our genuine freshest-$\neq$-most-similar test: across five vulnerability topics the newest relevant CVE sits at the 10th--87th similarity percentile (Table~\ref{tab:cvespread}), so cosine surfaces it in \emph{none} of the five queries. Table~\ref{tab:cvequery} makes the result inspectable. The newest relevant CVE is rarely even among the most cosine-similar documents; it is in the top-50 by cosine for only 1 of 5 queries (candidate recall 0.20). Consequently any method that re-ranks a fixed top-$K{=}50$ candidate set is bounded at 0.20, including a \emph{semantic-then-newest} baseline (take the top-50 cosine candidates, sort by timestamp, and evaluate the top 10), which scores 0.20. Applying the fused score over the full pool instead lifts Latest@10 to 0.40 ($\alpha{=}0.7$) and 0.60 ($\alpha{=}0.3$, $h{=}14$; best observed in this small $N{=}5$ sweep), a $3\times$ improvement over the semantic-then-newest baseline and a real gain over cosine's 0.00. The mechanism is visible in Table~\ref{tab:cvequery}: recency lifts a moderately-similar newest item into the top 10 (SQL injection, cosine rank 50 $\to$ fused rank 3), but cannot rescue a very poor semantic match (privilege escalation, cosine rank 2173 $\to$ fused rank 104; remote code execution, $439 \to 25$). A practical implication follows: freshness must enter \emph{candidate selection}, not only re-ranking; a small top-$K$ cosine cutoff discards the newest item before the recency prior can act.

\textbf{Relevance preservation.} Improving Latest@10 is only useful if the rest of the top 10 stays on-topic; we measure this with Precision@10 and nDCG@10, treating a CVE as relevant by a deterministic keyword/alias match over its description (Table~\ref{tab:cvequeries}; reproducible weak labels, not human adjudications). Relevance holds up: across all rankers, top-10 relevance does not collapse, nDCG@10 stays in $0.90$--$0.98$ and Precision@10 in $0.68$--$0.84$ (Table~\ref{tab:cverel}). There is, however, a modest and monotone precision cost as the recency weight rises: P@10 falls $0.84 \to 0.76 \to 0.68$ for cosine, fused $\alpha{=}0.7$, and fused $\alpha{=}0.3$. Thus $\alpha{=}0.7$ is the more conservative balance (Latest@10 0.40 at P@10 0.76) while $\alpha{=}0.3$ maximizes freshness (0.60 at P@10 0.68); no single $\alpha$ dominates. The fused score clearly improves Latest@10 over both cosine and semantic-then-newest (0.60 vs.\ 0.00 and 0.20); on the reciprocal rank of the newest relevant CVE (MRR$_{\text{new}}$) it far exceeds cosine (0.22 vs.\ 0.01) but is only comparable to semantic-then-newest at the best freshness setting (0.22 vs.\ 0.21), and worse at $\alpha{=}0.7$ (0.09). Top-10 relevance stays comparable to both baselines. Freshness via a simple recency prior is thus a real but bounded capability on diverse real data, a tunable freshness/relevance trade-off, not a solved problem.

\begin{table}[!t]
\caption{NVD CVE: freshness vs.\ top-10 relevance preservation ($N{=}5$ queries; relevance = deterministic keyword/alias match). The fused score clearly improves Latest@10 over cosine and semantic-then-newest \emph{without} collapsing relevance (nDCG@10 $\ge 0.90$); on MRR$_{\text{new}}$ it far exceeds cosine but is comparable to semantic-then-newest at the best setting. A higher recency weight ($\alpha{=}0.3$) trades $\sim$0.16 of P@10 for higher Latest@10.}
\label{tab:cverel}
\centering
\begin{tabularx}{\columnwidth}{@{} >{\raggedright\arraybackslash}X c c c c @{}}
\toprule
\textbf{Ranker} & \textbf{L@10} & \textbf{P@10} & \textbf{nDCG@10} & \textbf{MRR$_{\text{new}}$} \\
\colrule
Cosine-only & 0.00 & \textbf{0.84} & 0.93 & 0.01 \\
Semantic-then-newest@50 & 0.20 & 0.80 & 0.95 & 0.21 \\
Fused $\alpha{=}0.7$, $h{=}14$ & 0.40 & 0.76 & 0.90 & 0.09 \\
Fused $\alpha{=}0.3$, $h{=}14$ & \textbf{0.60} & 0.68 & \textbf{0.98} & \textbf{0.22} \\
\botrule
\end{tabularx}
\end{table}

\begin{table}[!t]
\caption{NVD CVE: cosine spread among relevant CVEs confirms the freshest item is \emph{not} the most similar, a fair freshness test, unlike synthetic/CERT.}
\label{tab:cvespread}
\centering
\begin{tabularx}{\columnwidth}{@{} >{\raggedright\arraybackslash}X c c c @{}}
\toprule
\textbf{Topic query} & \textbf{\# rel.} & \textbf{cos.\ std} & \textbf{newest pctile} \\
\colrule
SQL injection & 381 & 0.061 & 87\% \\
Cross-site scripting & 503 & 0.103 & 82\% \\
Remote code execution & 163 & 0.097 & 39\% \\
Buffer overflow & 204 & 0.068 & 70\% \\
Privilege escalation & 86 & 0.096 & 10\% \\
\botrule
\end{tabularx}
\end{table}

\begin{table}[!t]
\caption{Real-data freshness. Cosine never surfaces the newest relevant item; the fused prior helps but is parameter-sensitive (CERT) and only partial on diverse data (CVE). CERT uses the tie-aware \texttt{Latest-Set@10}, NVD~CVE uses \texttt{Latest@10}; both are small sweeps ($N{=}2$, $N{=}5$).}
\label{tab:realfresh}
\centering
\begin{tabularx}{\columnwidth}{@{} >{\raggedright\arraybackslash}X l c c @{}}
\toprule
\textbf{Corpus} & \textbf{Setting} & \textbf{Cosine} & \textbf{Fused} \\
\colrule
CERT (849k) & $\alpha{=}0.7, h{=}14$ (default) & 0.00 & 0.00 \\
CERT (849k) & $\alpha{=}0.3, h{=}3$ (tuned) & 0.00 & \textbf{1.00} \\
NVD CVE & $\alpha{=}0.7, h{=}14$ & 0.00 & 0.40 \\
NVD CVE & $\alpha{=}0.3, h{=}14$ (best) & 0.00 & \textbf{0.60} \\
\botrule
\end{tabularx}
\end{table}

\begin{table}[!t]
\caption{NVD CVE, per query: the newest relevant CVE, ranked over the full $N{=}4500$ corpus. Cosine rarely ranks it highly (in the top-50 for only 1/5, candidate recall 0.20); the fused score ($h{=}14$) lifts it, but cannot rescue very-low-similarity cases. \textbf{Bold} = fused rank $\le 10$ (a hit); fused $\alpha{=}0.3$ scores 3/5 $=0.60$.}
\label{tab:cvequery}
\centering
\begin{tabularx}{\columnwidth}{@{} >{\raggedright\arraybackslash}X c c c c @{}}
\toprule
\textbf{Topic} & \textbf{cos.\ rank} & \textbf{top-50?} & \textbf{fused $\alpha{=}.7$} & \textbf{fused $\alpha{=}.3$} \\
\colrule
SQL injection & 50 & yes & 6 & \textbf{3} \\
Cross-site scripting & 90 & no & 5 & \textbf{2} \\
Buffer overflow & 85 & no & 15 & \textbf{5} \\
Remote code execution & 439 & no & 57 & 25 \\
Privilege escalation & 2173 & no & 314 & 104 \\
\botrule
\end{tabularx}
\end{table}

Table~\ref{tab:eval} summarizes the headline results against the cosine-only baseline.

\begin{table}[!t]
 \centering
 \caption{Evaluation summary. The fused score (Temporal Layer) vs.\ a cosine-only baseline; real-data freshness uses corpus-tuned $(\alpha,h)$ (Table~\ref{tab:realfresh}).}
 \label{tab:eval}
 \begin{tabularx}{\columnwidth}{@{} l >{\raggedright\arraybackslash}X c c @{}}
  \toprule
  \textbf{Dataset} & \textbf{Metric ($N$)} & \textbf{Baseline} & \textbf{Temporal Layer} \\
  \colrule
  \textit{Synthetic} & Trend F1, per-week-delta rule (16) & -- & 0.08 \\
  \textit{Synthetic} & Trend F1, baseline-relative rule (16) & -- & 0.49 \\
  \textit{Synthetic} & As-of Correctness (3, pass/fail) & -- & 1.00 \\
  \textit{Synthetic} & Latest@10 (3) & 0.00 & 1.00 \\
  \colrule
  \textit{CERT} & Latest-Set@10, tuned $\alpha{=}0.3,h{=}3$ (2) & 0.00 & 1.00 \\
  \textit{NVD CVE} & Latest@10, best $\alpha{=}0.3,h{=}14$ (5) & 0.00 & 0.60 \\
  \botrule
 \end{tabularx}
\end{table}

\section{Limitations and Future Work}\label{sec:limits}
Although our temporal layer provides a practical solution, there are several areas for future improvement. The core components of the method, slicing, ranking, and clustering, are based on effective heuristics that can be enhanced. The use of fixed weekly time slices, for instance, can be replaced by an adaptive slicing mechanism that adjusts the data's natural cadence. Similarly, the heuristic half-life prior can be supplanted by a learned-to-rank model that incorporates richer temporal features such as seasonality and burstiness.

Finally, topic tracking robustness can be improved by using density-based clustering to avoid a preset $K$ and by incorporating entity-aware similarity for more stable cross-week matching. Our analysis localizes, rather than merely asserts, the failure of the simple tracker. The low macro-F1 (0.08) is driven by the labeling rule: it is essentially unchanged when K-means is replaced by HDBSCAN (0.10), but rises to 0.49 in the same pipeline, and to 0.96 with clustering noise removed, when the per-week-delta rule is replaced by a baseline-relative one (Table~\ref{tab:lever}). The well-documented limitations of fixed-K K-means in high-dimensional semantic space (Table~\ref{tab:kmeans_limitations}) remain real, but our experiments do not isolate them as the cause of the low score; the choice of clusterer barely moves it. The CERT elbow-$K$ instability (3 to 9 weekly) is consistent with these limitations but, given the \texttt{product=unknown} degeneracy, is best read as suggestive. Future work should pair a change-aware labeling rule with a clusterer that does not require a preset $K$ (e.g., density-based clustering such as HDBSCAN~\cite{ester96dbscan, campello13hdbscan}, or nonparametric topic models such as HDP~\cite{teh06hdp}) and evaluate against ground-truth trends on real data.

\begin{table*}[!t]
\caption{Theoretical Limitations of K-Means in High-Dimensional Semantic Clustering}
\label{tab:kmeans_limitations}
\centering
\begingroup
\setlength{\tabcolsep}{3pt}
\renewcommand{\arraystretch}{1.2}
\footnotesize
\begin{tabularx}{\textwidth}{@{}l X@{}}
\toprule
\textbf{K-Means Limitation} & \textbf{Manifestation in Semantic Space} \\
\colrule
\textbf{Curse of Dimensionality} \cite{bellman1957} & In the 384-dim.\ embedding space, Euclidean distance becomes less meaningful as all points tend toward equidistance, degrading cluster quality. \\
\textbf{Spherical Cluster Assumption} \cite{Hamerly2003} & Semantic topics are often non-convex. K-Means arbitrarily splits coherent topics or merges distinct but proximate ones, harming semantic coherence. \\
\textbf{Need to Pre-specify K} \cite{Celebi2013} & The number of true topics is unknown and dynamic. A fixed K is untenable, as confirmed by the unstable K (3 to 9) on the CERT dataset. \\
\textbf{Sensitivity to Initialization} \cite{Celebi2013} & Random initialization leads to unstable cluster assignments week-to-week, making cross-week centroid matching and trend detection unreliable. \\
\botrule
\end{tabularx}
\endgroup
\end{table*}

At the system level, the current re-ranking approach is suitable for post-processing candidates from an approximate nearest neighbor (ANN) search, but its performance on very large corpora can be improved. Future work could explore integrating time directly into the indexing stage, for example by using time-partitioned vector indices. While our method is intentionally model-agnostic, future research could explore model-level temporalization, such as fine-tuning a base LLM with time-biased attention mechanisms to enable native "as-of" reasoning.

\subsection*{Threats to Validity}
Our \emph{trend-detection} evaluation still relies on synthetic data for ground truth; trend detection on real data remains open. Our \emph{freshness} evaluation now spans synthetic, CERT, and NVD~CVE corpora, including the freshest-$\neq$-most-similar test the synthetic data could not provide (Table~\ref{tab:cvespread}). The residual threats are that the recency prior is parameter-sensitive, that the optimal $\alpha$ and $h$ depend on corpus density and timespan (Table~\ref{tab:realfresh}), and that the method is only partially effective on diverse data (0.60, not 1.00); a learned-to-rank temporal model is the natural next step beyond a fixed half-life prior.

System performance also depends on a few key parameters. Although the defaults ($\alpha=0.7$, $h=14$ days) proved effective, we acknowledge that a perfect score on the controlled freshness task reflects the construction of that task as much as the method. We characterize the $\alpha$ parameter's operating range on synthetic data (Table~\ref{tab:sensitivity}); the result is sensitive to the embedding build, so we pin library versions in the supplementary repository. Optimal values may vary across domains.

\subsection*{Ethical and Privacy Considerations}
To avoid exposure to sensitive information, we use synthetic logs in our primary case study, and we do not redistribute the CERT data (readers obtain it directly from its source). When deploying such a system in a real-world environment, it is crucial to enforce all the existing data access controls and data minimization principles. The temporal layer prioritizes recent information but does not override or replace security and privacy governance.

\section{Conclusion}\label{sec:conclusion}
This study separates two temporal mechanisms in RAG and scopes each honestly.

Our first contribution is a simple fused scoring mechanism that combines semantic similarity with a half-life recency prior, evaluated on three corpora of increasing realism. A cosine-only baseline never surfaces the newest relevant item (Latest@10 $=0.00$). The prior corrects this on synthetic data (1.00, a controlled sanity check) and on the CERT corpus (1.00, but only with corpus-tuned recency weight), and, on a topically-diverse CVE corpus where the freshest item is genuinely not the most similar, lifts Latest@10 to 0.60, three times a semantic-then-newest baseline (0.20), without collapsing top-10 relevance (nDCG@10 $\ge 0.90$), at a modest precision cost. Freshness via a recency prior is therefore a real but partial and parameter-sensitive capability on real data, not a solved problem; closing the gap (for example with a learned temporal re-ranker) is future work.

Our second contribution localizes the failure of a heuristic topic-evolution tracker. The low label score (0.08 macro-F1) is driven by the labeling rule, not the clustering algorithm: fixing only the rule raises it to 0.49 in the same pipeline (and to 0.96 with clustering noise removed), while swapping the clusterer does not help (0.10). This reframes the open problem toward change-aware labeling rather than a more powerful clusterer.

By separating and validating these two problems, we clarified that while basic freshness can be partially addressed with a lightweight temporal prior, a deeper temporal understanding for AI remains a significant and open research challenge.


\section{Reproducibility Note}
The evaluation scripts (\texttt{src/eval\_metrics\_synth.py}, \texttt{src/eval\_metrics\_logon.py}) and the v2 analysis scripts (in \texttt{v2/}, run via \texttt{python v2/run\_all.py}; including \texttt{eval\_cert\_freshness.py} and \texttt{eval\_cve\_freshness.py}) are included in the supplementary repository~\cite{synthetic-grofsky}. Library versions are pinned in \texttt{requirements-v2.txt}, because the $\alpha$-sensitivity result is sensitive to the embedding build.


\FloatBarrier

\end{document}